\documentclass{article}
\usepackage{spconf,amsmath,graphicx,hyperref}
\usepackage{cite}
\usepackage{amssymb}
\usepackage{multirow}
\usepackage{makecell}

\title{ENDO-G\textsuperscript{2}T: GEOMETRY-GUIDED \& TEMPORALLY AWARE TIME-EMBEDDED 4DGS FOR ENDOSCOPIC SCENES}
\name{Yangle Liu$^{1}$
\qquad Fengze Li$^{1,2}$
\qquad Kan Liu$^{1,2}$
\qquad Jieming Ma$^{2\ast}$
\thanks{$\ast$ Corresponding author.}}
\address{$^{1}$ University of Liverpool\\
$^{2}$ Xi'an Jiaotong--Liverpool University}
\begin{document}
%
\maketitle
\begin{abstract}
Endoscopic (endo) video exhibits strong view-dependent effects such as specularities, wet reflections, and occlusions. Pure photometric supervision misaligns with geometry and triggers early geometric drift, where erroneous shapes are reinforced during densification and become hard to correct. We ask how to anchor geometry early for 4D Gaussian splatting (4DGS) while maintaining temporal consistency and efficiency in dynamic endoscopic scenes. Thus, we present Endo-G\textsuperscript{2}T, a geometry-guided and temporally aware training scheme for time-embedded 4DGS. First, geo-guided prior distillation converts confidence-gated monocular depth into supervision with scale-invariant depth and depth-gradient losses, using a warm-up-to-cap schedule to inject priors softly and avoid early overfitting. Second, a time-embedded Gaussian field represents dynamics in XYZT with a rotor-like rotation parameterization, yielding temporally coherent geometry with lightweight regularization that favors smooth motion and crisp opacity boundaries. Third, keyframe-constrained streaming improves efficiency and long-horizon stability through keyframe-focused optimization under a max-points budget, while non-keyframes advance with lightweight updates. Across EndoNeRF and StereoMIS-P1 datasets, Endo-G\textsuperscript{2}T achieves state-of-the-art results among monocular reconstruction baselines.
\end{abstract}

\section{Introduction}
\label{sec:intro}
Endoscopic imaging is challenging for 3D and 4D reconstruction \cite{yang20243d}. Specularities and wet reflections break Lambertian assumptions; tissues deform nonrigidly with occasional topology changes; narrow field of view yields sparse baselines; and instruments cause frequent, structured occlusions. Classical endoscopic SLAM and multiview stereo use illumination models, shading/reflectance priors, tool masking, nonrigid registration, and stereo fusion \cite{grasa2013visual}, yet degrade under strong highlights, large elastic motion, low texture, and long sequences with drift \cite{zhou2024improved}. Failures include depth bias near glossy mucosa, instability at instrument boundaries, and cumulative pose error in extended procedures \cite{qiu2020endoscope}.

Neural rendering reframes reconstruction by learning radiance and geometry from images. NeRF methods model a continuous, view-dependent radiance field with high-fidelity novel views but require long optimization and dense sampling \cite{mildenhall2021nerf}. 3D Gaussian splatting (3DGS) \cite{3dgs} speeds up rendering by projecting Gaussian primitives to screen space with differentiable splatting, achieving real-time quality. Its dynamic extension, 4D Gaussian splatting (4DGS) \cite{wu20244d}, introduces time-varying primitives and a time-embedded motion space, enabling fast dynamic reconstruction and much faster convergence for videos.

Motivated by these advances, several works adapt neural rendering to endoscopy. EndoNeRF-style methods add tool-aware sampling, learned deformation fields, and stereo or photometric cues to regularize depth near tissue surfaces \cite{wang2022neuralendonerf,zha2023endosurf}. Endoscopic Gaussian-splatting variants, such as Endo-4DGS and successors, initialize from monocular pseudo-depth, use confidence-guided learning with surface-normal and depth regularization, and employ lightweight deformation heads for temporal modeling \cite{huang2024endo4dgs,li2025realstendogs}. Despite these gains, two issues persist: early geometric drift when pseudo-depth is biased or injected too strongly, which is then reinforced during densification; and temporal decoherence with uncontrolled point growth on long, fast, and occluded sequences, which reduces stability and frame rate.

These limitations imply two requirements: an appearance-agnostic geometric anchor early in training, and a schedule that preserves temporal consistency and efficiency. Recent visual-geometry transformers satisfy the first need by producing pixel-aligned monocular depth with confidence in a single pass, well suited for early distillation \cite{wang2025vggt}. Their streaming variant adds causal attention and cached memory for incremental inference over long sequences, inspiring our system design \cite{zhuo2025streaming}. On the representation side, higher-dimensional Gaussian fields with rotor-based rotations yield smooth, stable trajectories in time-embedded spaces \cite{duan20244drotor}, and streaming Gaussian pipelines show that keyframe selection and point-budget control effectively curb long-horizon drift while sustaining throughput \cite{yan2025instantIGS}.

Motivated by these observations, we propose ENDO-G\textsuperscript{2}T, a geometry guided and temporally aware training scheme for time-embedded 4DGS. Our design directly targets the above failure modes with three modules that align with the identified requirements. 

Thus, the key contributions are as follows. \textbf{(i)} ENDO-G\textsuperscript{2}T introduces geo-guided prior distillation (GPD) that anchors early geometry via confidence-gated monocular priors with scale-invariant and depth-gradient supervision under a warm-up-to-cap schedule; \textbf{(ii)} it employs a time-embedded gaussian field (TEGF) with a rotor-like rotation parameterization and lightweight regularization to improve temporal coherence and reduce floaters; \textbf{(iii)} it adopts keyframe-constrained streaming (KCS) with a max-points budget to curb point growth, sustain throughput, and mitigate long-horizon drift; \textbf{(iv)} across EndoNeRF and StereoMIS-P1 datasets, ENDO-G\textsuperscript{2}T achieves state-of-the-art monocular 4DGS results at high frame rates.
\begin{figure}
    \centering
    \includegraphics[width=1\columnwidth]{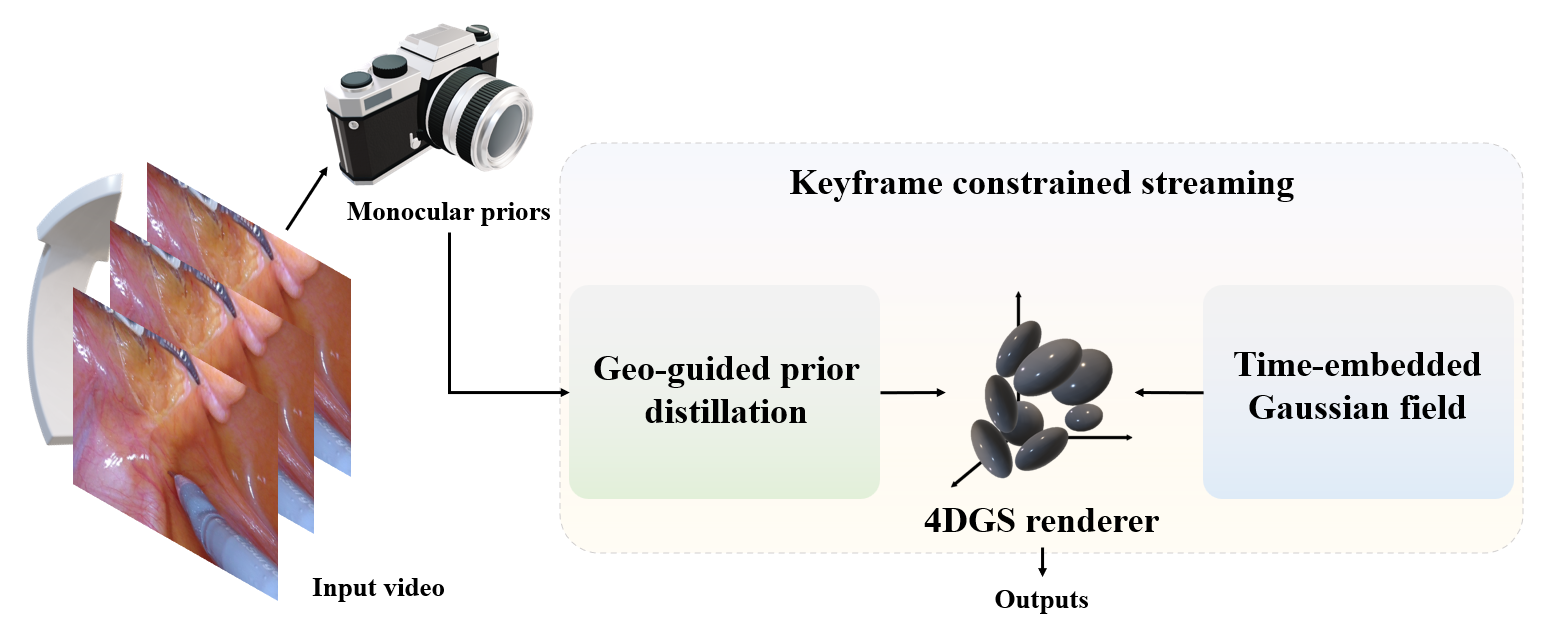}
    \caption{Overview of ENDO-G\textsuperscript{2}T.}
\end{figure}

\section{Endo-G\textsuperscript{2}T}
ENDO-G\textsuperscript{2}T comprises three modules: GPD that anchors geometry early with confidence-gated monocular depth under a warm-up-to-cap schedule, a TEGF that models dynamics in XYZT with a rotor-like rotation and lightweight regularization, and keyframe constrained streaming that preserves efficiency and long-horizon stability via keyframe-focused optimization and a global max-points budget. 

\subsection{Geo-guided prior distillation}

This module injects appearance-agnostic supervision early by distilling pixel-aligned monocular geometry into the rendered depth. For each training view we use the observed RGB image $I$, the current rendering $\hat I$ and rendered depth $\hat D$, together with external priors $(D^\star, C^\star)$ where $D^\star$ is the monocular depth and $C^\star\in[0,1]$ is its confidence. Supervision is restricted to a valid pixel set selected by (i) a frame-adaptive confidence threshold and (ii) a physically plausible depth range; optional instrument masks further exclude tool regions. Prior-based losses are activated only when the fraction of valid pixels is sufficiently large (at least 10\%) to avoid spurious supervision. Depths are normalized per frame and compared in the log domain to achieve scale invariance, and a depth-gradient term aligns edge structure. Both prior terms are introduced with a warm-up-to-cap schedule so that the influence of priors grows gradually and remains bounded, mitigating early overfitting to noisy estimates.

Photometric reconstruction combines an $L_1$ term and SSIM,
\begin{equation}
\label{eq:photo}
\begin{aligned}
\mathcal{L}_{\mathrm{photo}}
&= (1-\lambda_{\mathrm{dssim}})\,\|I-\hat I\|_1 \\
&\quad + \lambda_{\mathrm{dssim}}\big(1-\mathrm{SSIM}(I,\hat I)\big).
\end{aligned}
\end{equation}
where $\lambda_{\mathrm{dssim}}\in[0,1]$ balances the two components and SSIM is computed in a standard patchwise manner. We also define the valid pixel set by confidence, depth range, and an optional instrument mask:
\begin{equation}
\label{eq:validset}
\begin{aligned}
\Omega_{\mathrm v}
&= \Big\{\,p\ \text{pixel}\ :\ C^\star(p)\ge \tau,\ D_{\min}\le D^\star(p)\le D_{\max},\\
&\qquad\qquad\ \ \mathcal M_{\mathrm{inst}}(p)=0\,\Big\}.
\end{aligned}
\end{equation}
where $\tau=\max\{0.01,\ 0.5\,\max_{q} C^\star(q)\}$ is an adaptive confidence threshold, $D_{\min},D_{\max}$ bound plausible depths, and $\mathcal M_{\mathrm{inst}}\in\{0,1\}^{H\times W}$ is an optional instrument mask. Prior-based terms are enabled only if $|\Omega_{\mathrm v}|/(HW)\ge 0.1$.

To compare geometry without enforcing an absolute scale, depths are min to max normalized per frame in code. We then measure a scale invariant discrepancy in the log domain on $\Omega_{\mathrm{v}}$,
\begin{equation}
\label{eq:silog}
\mathcal{L}_{\mathrm{SILog}}
= 10\,\sqrt{\mathrm{Var}_{p\in\Omega_{\mathrm{v}}}\!\big(g(p)\big)
+ \beta\,\mathrm{Mean}_{p\in\Omega_{\mathrm{v}}}\!\big(g(p)\big)^2 }.
\end{equation}

\begin{equation}
\label{eq:silog_res}
g(p)=\log\!\big(\tilde D(p)+\epsilon\big)-\log\!\big(\tilde D^\star(p)+\epsilon\big).
\end{equation}
where $\tilde D$ and $\tilde D^\star$ are the per frame normalized versions of $\hat D$ and $D^\star$, $\epsilon>0$ prevents singularities, and $\beta>0$ controls the penalty on global log scale bias. To sharpen geometry at boundaries we align first order depth gradients on the valid set,
\begin{equation}
\label{eq:grad}
\begin{aligned}
\mathcal{L}_{\mathrm{grad}}
&= \frac{1}{|\Omega_{\mathrm{v}}|}\sum_{p\in\Omega_{\mathrm{v}}}
\Big(\|\nabla_x \hat D(p)-\nabla_x D^\star(p)\|_1 \\
&\qquad\qquad\quad + \|\nabla_y \hat D(p)-\nabla_y D^\star(p)\|_1\Big).
\end{aligned}
\end{equation}
where $\nabla_x$ and $\nabla_y$ are forward differences along the horizontal and vertical axes in pixels.

To avoid overfitting to prior noise at the beginning of training, the prior weights follow a warm up to cap schedule,
\begin{equation}
\label{eq:warmcap}
\begin{aligned}
\lambda_{\mathrm{SI}}(t)
&= \lambda_{\mathrm{SI},0}\,\min\!\Big(1,\frac{t}{T_{\mathrm{warm}}}\Big)\,w_{\max},\\
\lambda_{\nabla}(t)
&= \lambda_{\nabla,0}\,\min\!\Big(1,\frac{t}{T_{\mathrm{warm}}}\Big)\,w_{\max}.
\end{aligned}
\end{equation}
where $t$ is the global iteration, $T_{\mathrm{warm}}>0$ is the warm up length, $\lambda_{\mathrm{SI},0},\lambda_{\nabla,0}>0$ are base coefficients, and $w_{\max}\in(0,1]$ caps the effective strength of priors. The module objective is
\begin{equation}
\label{eq:geo}
\mathcal{L}_{\mathrm{geo}}=\mathcal{L}_{\mathrm{photo}}+\lambda_{\mathrm{SI}}(t)\,\mathcal{L}_{\mathrm{SILog}}+\lambda_{\nabla}(t)\,\mathcal{L}_{\mathrm{grad}}.
\end{equation}
All symbols in \eqref{eq:photo} to \eqref{eq:geo} are defined above and finite value checks are applied before accumulation to ensure numerical stability.

\subsection{Time-embedded Gaussian field}

This module represents dynamics by lifting Gaussian primitives into a time-embedded space and evolving their parameters smoothly. For each primitive we maintain a 3D center, a diagonal scale, a rotation, an opacity, and spherical-harmonic color coefficients. The covariance used for splatting is
\begin{equation}
\label{eq:cov}
\Sigma_i(t)=R_i(t)\,S_i^2(t)\,R_i^\top(t),
\end{equation}
where $\mu_i(t)\in\mathbb{R}^3$ is the center, $S_i(t)$ contains axis scales, $R_i(t)\in\mathrm{SO}(3)$ is the rotation, and thus $\Sigma_i(t)\in\mathbb{R}^{3\times 3}$ is positive definite by construction; $\alpha_i(t)\in(0,1)$ is opacity and $\mathbf{c}_i(t)\in\mathbb{R}^{(n+1)^2\times 3}$ holds spherical-harmonic coefficients up to degree $n$. Positions later advance with a per-primitive velocity and rotations are updated by a minimal rotor operator; the corresponding variables $v_i(t)$, $\rho_i(t)$ and the frame interval $\Delta t>0$ are introduced in the subsequent evolution equations.

To bias toward decisive visibility and coherent motion with minimal overhead we include two lightweight stabilizers. Opacity entropy encourages $\alpha_i(t)$ to concentrate near zero or one,
\begin{equation}
\label{eq:entropy}
\begin{aligned}
\mathcal{L}_{\mathrm{ent}}
&= -\frac{1}{N}\sum_{i=1}^{N}\Big(\alpha_i(t)\log\alpha_i(t) \\
&\qquad\qquad + \big(1-\alpha_i(t)\big)\log\big(1-\alpha_i(t)\big)\Big).
\end{aligned}
\end{equation}
where $N$ is the number of active primitives at time $t$. Local velocity coherence smooths motion in a joint space and time neighborhood,
\begin{equation}
\label{eq:vel}
\begin{aligned}
\mathcal{L}_{\mathrm{vel}}
&= \frac{1}{N}\sum_{i=1}^{N}\frac{1}{|\mathcal{N}_k(i,t)|}
   \sum_{j\in\mathcal{N}_k(i,t)}
   \Big\|
   \underbrace{\mu_i(t)-\mu_i(t-\Delta t)}_{v_i(t)} \\
&\qquad\qquad\qquad\quad
   - \underbrace{\mu_j(t)-\mu_j(t-\Delta t)}_{v_j(t)}
   \Big\|_1 .
\end{aligned}
\end{equation}
where $\mathcal N_k(i,t)$ denotes the $k$ nearest neighbors of primitive $i$ in a joint position–time metric, where $k\in\mathbb{N}$ is a small fixed neighborhood size. When $N$ is large, the inner sum is evaluated on a subsample to respect memory limits. All symbols in \eqref{eq:cov} to \eqref{eq:vel} are defined above.

\subsection{Keyframe constrained streaming}
This module maintains stability and throughput on long sequences by interleaving full refinement with lightweight updates. The video is partitioned by a stride $w\in\mathbb{N}$ into keyframes $\mathcal{K}$ and candidate frames $\mathcal{C}$, where
\begin{equation}
\label{eq:kset}
\begin{aligned}
\mathcal{K} &= \{\,f\in\{1,\dots,F\}:\ f\equiv 1\ \mathrm{mod}\ w\,\},\\
\mathcal{C} &= \{1,\dots,F\}\setminus\mathcal{K}.
\end{aligned}
\end{equation}

Keyframes run full optimization with densification and pruning, while candidate frames apply lightweight image-space updates that keep the model size fixed. To prevent uncontrolled growth, the active set of Gaussians obeys a global budget at every time $t$:
\begin{equation}
\label{eq:budget}
\begin{aligned}
|\mathcal{G}_t| &\le G_{\max},\\
&\text{for all } t \text{ associated with frames } 1,\dots,F.
\end{aligned}
\end{equation}

At keyframes the budget is enforced by balancing additions and removals, which curbs long-horizon drift and preserves frame rate. For throughput reporting, a raster-only mode disables input and output so that measured frames per second reflect GPU rasterization alone.

\begin{figure}
    \centering
    \includegraphics[width=1\columnwidth]{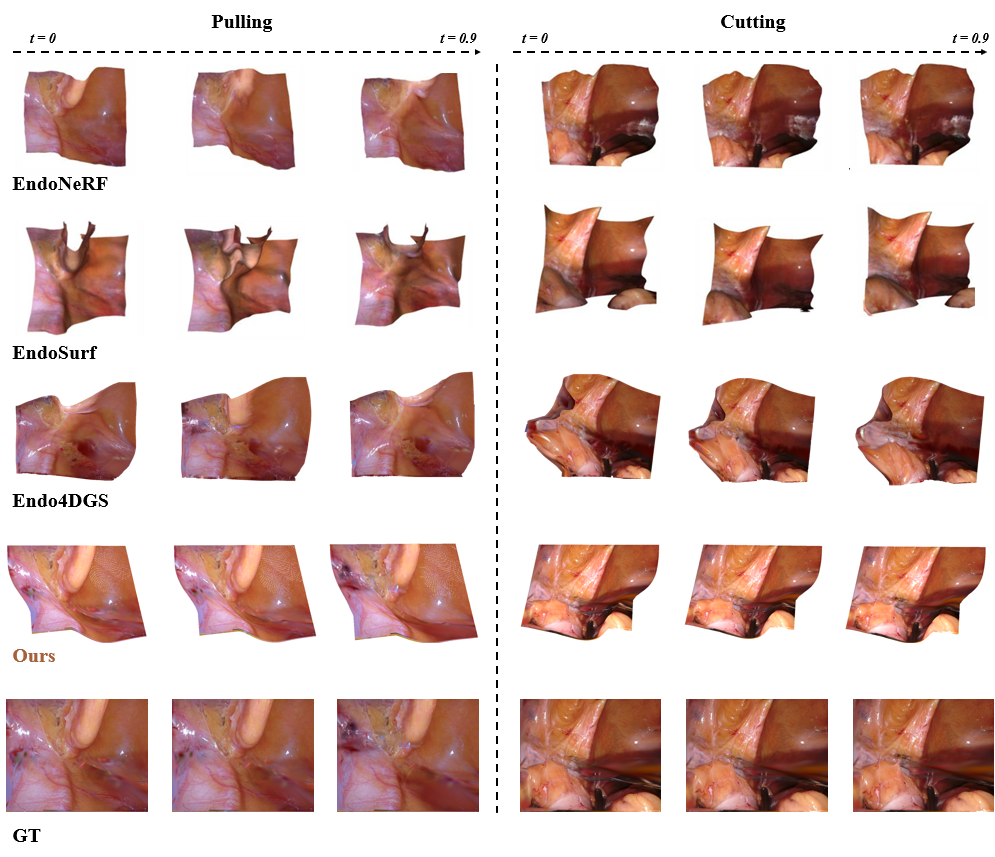}
    \caption{Comparison on EndoNeRF Cutting/Pulling: Endo-G\textsuperscript{2}T produces sharper tissue boundaries and fewer floaters than other baselines.}
    \label{fig:result1}
\end{figure}

\begin{figure}
    \centering
    \includegraphics[width=1\columnwidth]{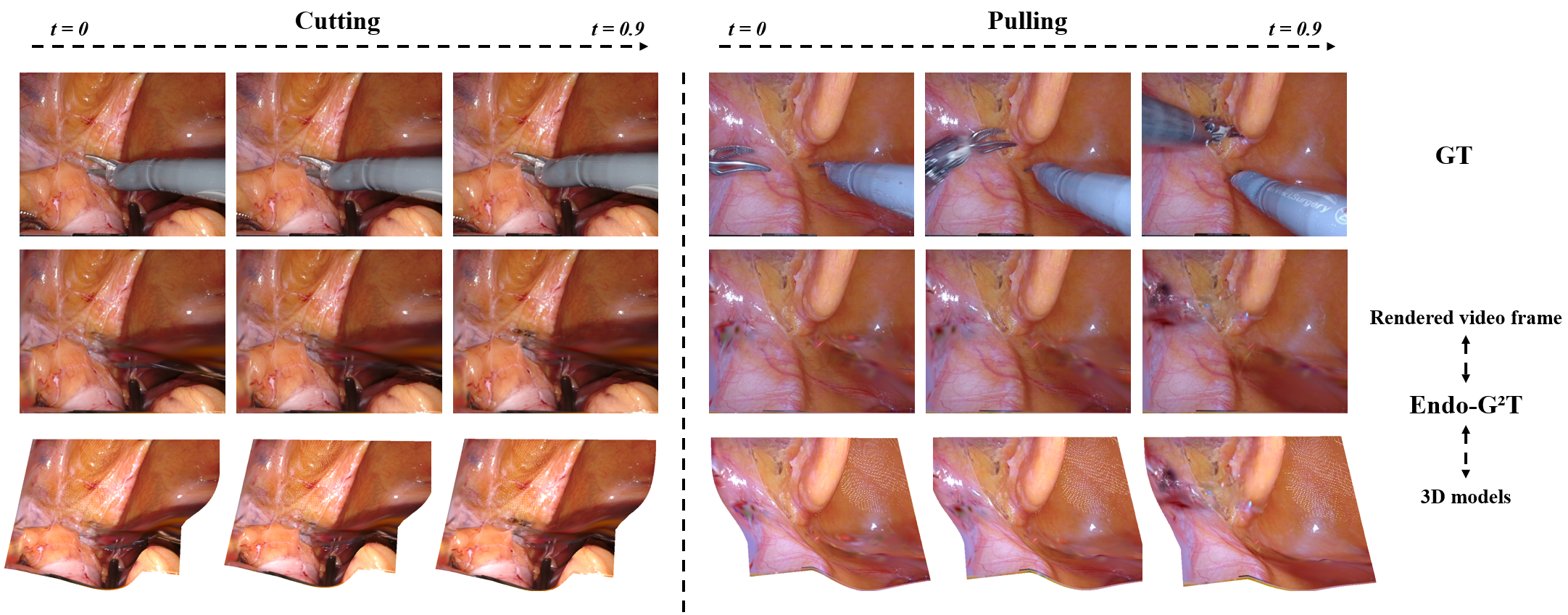}
    \caption{Endo-G\textsuperscript{2}T on EndoNeRF Cutting/Pulling from $t{=}0$ to $t{=}0.9$: GT, our renders, and reconstructed surfaces, showing coherent geometry and crisp opacity.}
    \label{fig:result2}
\end{figure}

\begin{table*}[t]
\caption{Quantitative comparison on EndoNeRF (cutting, pulling) and StereoMIS-P1 (frames 800–1000). Best results are in bold. FPS is measured in raster-only mode.}
\centering
\resizebox{\textwidth}{!}{
\begin{tabular}{l|ccc|ccc|ccc|c}
\noalign{\smallskip}\hline
\multirow{2}{*}{Models}
& \multicolumn{3}{c|}{EndoNeRF–Cutting} 
& \multicolumn{3}{c|}{EndoNeRF–Pulling}
& \multicolumn{3}{c|}{StereoMIS–P1 (800–1000)}
& \multirow{2}{*}{FPS $\uparrow$} \\ \cline{2-10}
& PSNR $\uparrow$ & SSIM $\uparrow$ & LPIPS $\downarrow$
& PSNR $\uparrow$ & SSIM $\uparrow$ & LPIPS $\downarrow$
& PSNR $\uparrow$ & SSIM $\uparrow$ & LPIPS $\downarrow$
& \\ \hline
EndoNeRF~\cite{wang2022neuralendonerf} 
& 35.840 & 0.942 & 0.057 
& 35.430 & 0.939 & 0.064 
& 30.500 & 0.880 & 0.078 
& 0.2 \\
EndoSurf~\cite{zha2023endosurf} 
& 34.893 & 0.952 & 0.107 
& 34.910 & 0.955 & 0.124 
& 31.221 & 0.890 & 0.071 
& 0.04 \\
Endo-4DGS~\cite{huang2024endo4dgs} 
& 36.165 & 0.959 & 0.039 
& 37.014 & 0.960 & 0.041 
& 32.188 & 0.898 & 0.066 
& 100 \\
ST-Endo4DGS~\cite{li2025realstendogs}
& 39.290 & 0.973 & 0.016
& 38.280 & 0.966 & 0.024
& 32.900 & 0.905 & 0.060
& 123 \\ \hline
\textbf{Endo-G\textsuperscript{2}T (ours)}
& \textbf{40.080} & \textbf{0.982} & \textbf{0.007}
& \textbf{38.290} & \textbf{0.970} & \textbf{0.016}
& \textbf{33.580} & \textbf{0.914} & \textbf{0.056}
& \textbf{148} \\\hline
\end{tabular}}
\label{tab:comparison}
\end{table*}

\begin{table*}[t]
\caption{Ablation on KCS. “KF” denotes keyframe scheduling, “w” is the keyframe stride, “KF-only” optimizes keyframes with candidates frozen, “w/o KF (w=1)” disables keyframe scheduling, and “budget” is the global max-points cap $G_{\max}$.}
\centering
\resizebox{\textwidth}{!}{
\begin{tabular}{l|ccc|ccc|ccc}
\noalign{\smallskip}\hline
\multirow{2}{*}{Variants}
& \multicolumn{3}{c|}{EndoNeRF--Cutting} 
& \multicolumn{3}{c|}{EndoNeRF--Pulling}
& \multicolumn{3}{c}{StereoMIS--P1 (800--1000)} \\ \cline{2-10}
& PSNR $\uparrow$ & SSIM $\uparrow$ & LPIPS $\downarrow$
& PSNR $\uparrow$ & SSIM $\uparrow$ & LPIPS $\downarrow$
& PSNR $\uparrow$ & SSIM $\uparrow$ & LPIPS $\downarrow$ \\ \hline
Endo-G\textsuperscript{2}T (w/ KF, w/o budget) 
& 38.912 & 0.971 & 0.012 
& 37.041 & 0.962 & 0.020 
& 32.217 & 0.902 & 0.061 \\
Endo-G\textsuperscript{2}T (KF-only, $w{=}5$) 
& 39.146 & 0.975 & 0.010 
& 37.498 & 0.965 & 0.019 
& 32.803 & 0.907 & 0.059 \\
Endo-G\textsuperscript{2}T (w/o KF, $w{=}1$) 
& 39.867 & 0.979 & 0.008 
& 38.061 & 0.968 & 0.016 
& 33.012 & 0.910 & 0.057 \\
Endo-G\textsuperscript{2}T (w/ KF, $w{=}10$) 
& 39.722 & 0.978 & 0.009 
& 38.003 & 0.967 & 0.016 
& 33.164 & 0.911 & 0.057 \\
Endo-G\textsuperscript{2}T (w/ KF, $w{=}3$) 
& 39.954 & 0.981 & 0.007 
& 38.241 & 0.970 & 0.016 
& 33.471 & 0.914 & 0.057 \\ \hline
\textbf{Endo-G\textsuperscript{2}T (w/ KCS, w/ budget, $w{=}5$)} 
& \textbf{40.080} & \textbf{0.982} & \textbf{0.007}
& \textbf{38.290} & \textbf{0.970} & \textbf{0.016}
& \textbf{33.580} & \textbf{0.914} & \textbf{0.056} \\ \hline
\end{tabular}}
\label{tab:kcs_ablation_ordered}
\end{table*}

\section{Experiments}
We evaluate ENDO-G\textsuperscript{2}T on EndoNeRF (cutting, pulling)  \cite{wang2022neuralendonerf} and StereoMIS \cite{hayoz2023learningstereomis}. For EndoNeRF, we use a 7:1 train–validation split and report PSNR, SSIM, and LPIPS on held-out views. For StereoMIS, following the Endo-4DGS \cite{huang2024endo4dgs} protocol, we reconstruct frames 800–1000 from scene P1 in a monocular setting to ensure fair comparison. All models are trained in PyTorch with Adam on a single NVIDIA RTX 4090, using a learning rate of $1.6\times10^{-3}$, mixed photometric supervision (L1+SSIM), and our warm-up–to–cap geometry-prior schedule. Inference FPS is measured in a raster-only mode that disables I/O to reflect pure GPU rasterization.

Table~\ref{tab:comparison} and Figures~\ref{fig:result1}–\ref{fig:result2} jointly show quantitative and qualitative gains. On EndoNeRF–Cutting, Endo-G\textsuperscript{2}T achieves 40.080 PSNR, 0.982 SSIM, and 0.007 LPIPS, improving over ST-Endo4DGS by 0.790 PSNR (2.0\% relative), +0.009 SSIM, and a 56.3\% LPIPS reduction. On EndoNeRF–Pulling, our method reaches 38.290/0.970/0.016, surpassing Endo-4DGS by +3.45\% PSNR, +0.010 SSIM, and 61.0\% lower LPIPS. On StereoMIS–P1, we obtain 33.580/0.914/0.056 with a +2.1\% PSNR gain over ST-Endo4DGS and a 6.7\% LPIPS drop. The qualitative frames in Figure~\ref{fig:result1} show sharper tissue boundaries and fewer floaters, while Figure~\ref{fig:result2} illustrates temporally coherent geometry and crisp opacity across time. Raster-only throughput rises to 148 FPS, a 20.3\% increase over ST-Endo4DGS, indicating that accuracy gains come without sacrificing speed.

For ablations, the core modeling components (GPD and TEGF) remain fixed to ensure backbone fairness; we vary only the system schedule in Table~\ref{tab:kcs_ablation_ordered}. Removing the global budget while keeping keyframes degrades Cutting PSNR from 40.080 to 38.912 (2.9\%) and raises LPIPS from 0.007 to 0.012 (71\% higher), reflecting point explosion and instability. Freezing candidates (KF-only) saves compute but trails the full model across datasets, and disabling keyframes (w=1) recovers some accuracy yet loses the periodic re-anchoring benefits. Varying the stride reveals a sweet spot at $w{=}5$ yielding slightly lower PSNR/SSIM or higher LPIPS. The full KCS configuration (budget + $w{=}5$) is consistently best, supporting the proposed efficiency–stability trade-off.

\section{Conclusion}
We presented ENDO-G\textsuperscript{2}T, a geometry-guided and temporally aware training scheme for time-embedded 4D Gaussian splatting in endoscopy. The method unifies geo-guided prior distillation, a time-embedded Gaussian field with rotor-based evolution, and keyframe-constrained streaming with a max-points budget. Experiments on EndoNeRF and StereoMIS–P1 show state-of-the-art accuracy with high throughput, and ablations verify the importance of the scheduling strategy. Future work will study cross-view prior calibration, uncertainty-aware fusion, and adaptive keyframe selection for clinical deployment.


\bibliographystyle{IEEEbib}
\bibliography{strings,refs}

\end{document}